\documentclass[conference]{IEEEtran}
\IEEEoverridecommandlockouts

\usepackage{balance}
\usepackage{cite}
\usepackage{enumitem}
\usepackage{amsmath,amssymb,amsfonts}
\usepackage{algorithmic}
\usepackage{graphicx}
\usepackage{textcomp}
\usepackage{xcolor}
\usepackage[hidelinks]{hyperref}
\usepackage{threeparttable}
\def\BibTeX{{\rm B\kern-.05em{\sc i\kern-.025em b}\kern-.08em
    T\kern-.1667em\lower.7ex\hbox{E}\kern-.125emX}}
\begin{document}

\title{Beyond Architectures: Evaluating the Role of Contextual Embeddings in Detecting Bipolar Disorder on Social Media\thanks{DOI reference number: 10.18293/SEKE2025-083}
}

\author{
    \IEEEauthorblockN{Khalid Hasan}
    \IEEEauthorblockA{
        \textit{Department of Computer Science} \\
        \textit{Missouri State University}\\
        Springfield, MO, USA \\
        kh597s@missouristate.edu
    }
    \and
    \IEEEauthorblockN{Jamil Saquer}
    \IEEEauthorblockA{
        \textit{Department of Computer Science} \\
        \textit{Missouri State University}\\
        Springfield, MO, USA \\
        jamilsaquer@missouristate.edu
    }
 }

\maketitle

\begin{abstract}

Bipolar disorder is a chronic mental illness frequently underdiagnosed due to subtle early symptoms and social stigma.
This paper explores the advanced natural language processing (NLP) models for recognizing signs of bipolar disorder based on user-generated social media text. 
We conduct a comprehensive evaluation of transformer-based models (BERT, RoBERTa, ALBERT, ELECTRA, DistilBERT) and Long Short Term Memory (LSTM) models based on contextualized (BERT) and static (GloVe, Word2Vec) word embeddings. Experiments were performed on a large, annotated dataset of Reddit posts after confirming their validity through sentiment variance and judgmental analysis. 
Our results demonstrate that RoBERTa achieves the highest performance among transformer models with an F1 score of $\sim$98\% while LSTM models using BERT embeddings yield nearly identical results. In contrast, LSTMs trained on static embeddings fail to capture meaningful patterns, scoring near-zero F1. These findings underscore the critical role of contextual language modeling in detecting bipolar disorder.  In addition, we report model training times and highlight that DistilBERT offers an optimal balance between efficiency and accuracy. In general, our study offers actionable insights for model selection in mental health NLP applications and validates the potential of contextualized language models to support early bipolar disorder screening.
\end{abstract}


\section{Introduction}

Bipolar disorder (BD) is a severe mental illness characterized by recurring manic and depressive cycles, affecting mood, energy, and thought patterns~\cite{WorldHealth}. Approximately 40 million people  worldwide suffered from bipolar disorder in 2019~\cite{WorldHealth}, and it is a serious public health concern. Besides the intrapersonal suffering of mood swings, bipolar disorder is associated with high disability rates and elevated risk of suicide. Unfortunately, diagnosis is usually delayed: on average, individuals with bipolar disorder wait nearly a decade for an accurate diagnosis, frequently receiving misdiagnoses (e.g., as unipolar depression)~\cite{RCPsych}. During this prolonged lost decade, patients are likely to experience deteriorating symptoms and even life-threatening crises. A recent study found that approximately 34\% of individuals with undiagnosed bipolar disorder had attempted suicide while waiting for a proper diagnosis~\cite{RCPsych}. These statistics underscore the urgent need for early intervention. 

In the digital era, many turn to social media and online forums to express emotional struggles and seek mental health support. Social websites like Reddit have forums (e.g., r/bipolar) where individuals with personal histories of coping with bipolar disorder connect and exchange information. These user-generated posts offer a valuable opportunity: if we can correctly detect linguistic indicators of bipolar symptoms in these posts, we might be able to assist in flagging vulnerable individuals or initiating earlier clinical evaluation. 

Natural language processing (NLP) techniques have already shown promise for related tasks such as detecting depression and suicidal ideation from text~\cite{ezerceli2024mental}. Previous research has shown that deep learning models, such as the transformer-based BERT, perform better than typical machine learning practices in detecting mental health~\cite {zhang2022natural,ezerceli2024mental}. However, numerous issues have arisen in modeling such data properly: do we acquire more from more complex contextual embeddings or the highly complex architectural designs? Will less sophisticated models with good embeddings catch up to end-to-end transformers? Our objective in this study is to conduct a systematic, pragmatic comparison of existing advanced NLP models for the binary prediction of bipolar disorder.

In particular, we focus on two dimensions: embedding strategy (i.e., static vs. contextual text embeddings) and model framework (LSTM-based vs. Transformer-based classifiers). To support this study, we curated a large dataset of Reddit posts annotated for bipolar disorder markers (presence/absence). We trained and tested a range of classifiers: state-of-the-art transformers, and multiple LSTM/BiLSTM variants (with and without attention mechanisms), using different word embeddings (pre-trained BERT, GloVe, and Word2Vec). We used standardized training and testing protocols to ensure result comparability. Our findings provide actionable guidance for practitioners seeking an approach that balances performance against resource consumption. 

\textbf{In brief, our contributions are as follows:}

\begin{itemize}
    \item We establish a new dataset of social media posts for bipolar disorder detection and perform extensive experiments comparing state-of-the-art transformer models to LSTM-based models with diverse embeddings. The dataset is available to researchers upon request.

    \item We conduct an in-depth study of the embedding procedures: demonstrating that contextual BERT-based embeddings significantly outperform static word vectors in detecting mental health signals, and provide reasons why static embeddings fail to detect nuanced bipolar disorder indicators.
    
    \item We contrast the impact of model structure and design choices: transformer versus LSTM, bi-directional models, and attention mechanism. Our results reveal that a hybrid LSTM architecture with BERT embeddings achieves comparable performance to full transformer models, whereas LSTMs with fixed embeddings exhibit substantially inferior results.
    
    \item We evaluate empirical factors such as training time, demonstrating that lighter transformers (DistilBERT) offer faster training with minimal performance trade-offs compared to static embedding-based models requiring longer training times, yet yield poor outcomes.
\end{itemize}

\section{Related Work}

Early efforts in automatic mental health detection from text relied on classical machine learning and hand-crafted features. For example, researchers have applied support vector machines and random forests on lexical features to identify users with depression or suicidal risk~\cite{ezerceli2024mental}. These traditional approaches showed some success, but they often required extensive feature engineering and could miss subtle linguistic cues. As larger datasets of social media content became available and deep learning rose to prominence, attention shifted toward neural network models that learn features directly from text. Recurrent neural networks (RNNs), especially LSTMs, were popular in several studies for tasks like depression detection on Reddit~\cite{pirina-coltekin-2018-identifying} and suicidal risk assessment, often using traditional word embeddings (such as Word2Vec or GloVe) to represent text. Traditional embeddings offer general semantic representations for words, but embed words into static vectors independent of context. This limits the analysis of nuanced user expressions about mental health. Contextual embeddings, in contrast, construct word representations based on how a word is used. This makes them more effective in capturing the subtle expressions existing in user-generated content~\cite{devlin2019bert}. 

The advent of transformer-based language models like BERT has significantly impacted NLP in general, and mental disorder detection is no exception. A recent narrative review by Zhang et al. noted an upward trend in using NLP for mental illness detection, with deep learning methods outperforming earlier techniques~\cite{zhang2022natural}. In particular, large pre-trained models have shown state-of-the-art performance on tasks such as identifying signs of depression, PTSD, and suicidal ideation in text~\cite{pirina-coltekin-2018-identifying}. For instance, in the domain of suicidal ideation detection, Hasan et. al. compared BERT and its variants to LSTM models on Reddit data, finding that RoBERTa achieved over 93\% F1 while LSTM (augmented with BERT embeddings and an attention mechanism) was only slightly behind at ~92.7\% F1~\cite{hasan2024comparative}. Their work also underscored that LSTMs with static embeddings (GloVe, Word2Vec) performed dramatically worse, often 15–20\% lower in accuracy. These results align with a consensus: contextual embeddings from transformers capture mental health-related language more effectively than static embeddings. 

Few studies have focused specifically on bipolar disorder detection from text, as much of the prior work has targeted depression or suicidal ideation. Bipolar detection is challenging because the language signals may be more subtle or varied (manic vs. depressive tone), and there is less explicit mention (users might not always name their condition). Moreover, bipolar disorder usually contains symptoms that overlap with other mental health conditions such as depression and social anxiety~\cite{vieta2018bipolar}. One recent effort by Lee et al. explored a multi-task approach to detect bipolar disorder risk and mood swings from longitudinal social media data, combining transformer-based text modeling with time-series analysis~\cite{lee-etal-2024-detecting-bipolar}. They reported that transformer-based models outperformed non-contextual models for bipolar risk identification, reinforcing the value of contextual language understanding. Our work differs in focusing on a binary post-level classification and directly comparing a wide range of modeling strategies under a unified experimental setting. According to our knowledge, this is the first study to comprehensively evaluate state-of-the-art transformers and traditional LSTM architectures with various embeddings on a bipolar disorder detection task.  

\section{Problem Definition}

The primary objective of this research is to classify bipolar-related Reddit social media posts using transformer-based and LSTM-based models. The issue is framed as a binary classification problem to identify whether a post belongs to bipolar disorder.

\textbf{Problem Statement} - \textit{
For a labeled dataset D = \{$d_1$, $d_2$,., $d_n$\} of $n$ posts, develop text classification models trained to predict for each post $d_i$ a label $p_i \in \{0, 1\}$, where $0$ indicates the absence of bipolar disorder and $1$ indicates its presence.
}
 
To address this problem, we composed four research questions, each focused on evaluating the performance of classification models. These questions are as follows.

\begin{enumerate}
    \item How do transformer-based models perform in bipolar disorder detection compared to traditional LSTM-based architectures?

    \item To what extent does embedding selection (contextual vs. static) influence model performance?

    \item Does adding attention mechanisms to LSTM/BiLSTM architectures improve performance significantly?

    \item How does training time vary across different models and embedding strategies, and what are the practical trade-offs?
\end{enumerate}

\section{Methodology}

\begin{figure}
    \centering
    \includegraphics[width=1\linewidth]{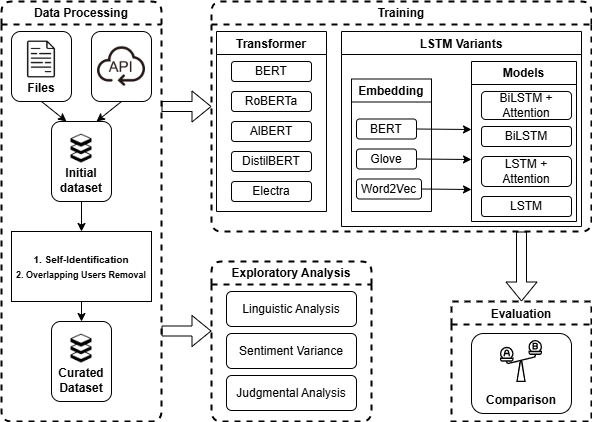}
    \caption{Summary of our Research Plan}
    \label{fig:research_architecture}
\end{figure}

Figure~\ref{fig:research_architecture} demonstrates an overview of our research plan.

\subsection{Data Collection and Annotation}

We collected abundant posts from the online discussion board platform Reddit to create a dataset to determine bipolar disorder. Using the Pushshift API\footnote{https://pushshift.io/}, we extracted posts from many subreddits during two years from January 2021 to December 2022, showcased in Table~\ref{tab:dataset_overview}. We identified r/bipolar as a primary subreddit where people post about living with bipolar. As such, posts from this subreddit serve as positive class instances (bipolar-relevant content). For the negative class, we gathered posts from a variety of other subreddits that are not about bipolar disorder. These were from general mental health and popular unrelated subreddits, such as r/depression, r/socialanxiety, r/TrueOffMyChest, r/confidence, and r/geopolitics. This approach uses a strategic mix of posts:

\begin{enumerate}[label=\arabic*., leftmargin=0pt, itemindent=15pt]
    \item Posts about other mental health issues (e.g., depression) to train the classifier to distinguish bipolar-specific content from similar disorders, creating a challenging but necessary distinction task.
    \item Completely unrelated posts to provide unambiguous negative examples.
\end{enumerate}

We selected these subreddits for a variety of reasons. Most importantly, we examined the total number of members within each group and their active membership because this would allow us to get significant, relevant information for model training. Moreover, we examined the subreddits with sufficient text information to label them into two labels: Bipolar if it originated from r/bipolar, and Non-bipolar if it originated from the other subreddits. We also carried out sentiment variance and statistical judgmental analysis as described in sections \ref{subsec:sentiment_variance} and \ref{subsec:judgmental_analysis} to verify our labeling. 

\subsection{Data Processing}

\begin{table}[htb]
    \caption{An overview of the Dataset}
    \centering
    \begin{tabular}{c|c|c}
        \hline
         & \textbf{Initial Count (K)} & \textbf{After Processed (K)} \\
         \hline
        \textbf{Bipolar} & 28K & 23K \\
        \hline
        \textbf{Non-bipolar} & 27K & 26K \\
    \hline
    \end{tabular}
    \label{tab:dataset_overview}
\end{table}

We used the self-identification techniques employed in mental health discourse analysis~\cite{coppersmith-etal-2015-adhd, mitchell-etal-2015-quantifying, kim2023understanding}. We created specific regular expressions to find the users who had directly written about their bipolar status. We also separated the common users who had posted within both the bipolar and non-bipolar subreddit communities to prevent data contamination. We followed the filtering strategy of Cohan et al.~\cite{cohan-etal-2018-smhd} in excluding the posts of these overlapping users to create a clean separation between the two categories.

We systematically preprocessed the raw text data for analysis, including tokenization, cleaning, normalization, and lemmatization. We began with whitespace tokenization, which tokenizes the text into words. Then, normalization used regular expressions to remove emojis, URLs, special characters, and numbers, retaining only significant linguistic features. Finally, lemmatization was utilized to reduce words to root form to enhance consistency and generalizability in representations within the text.

For various analysis purposes, we employed text data at different levels of preprocessing. For instance, for part-of-speech (POS) tagging and sentiment variance analysis, we used only tokenized text to avoid disrupting the original linguistic structure required for accurate tagging. Likewise, to identify mental health-related phrases, we utilized the original unprocessed text to maintain contextual detail and identify valuable multi-word expressions.

This systematic pre-processing technique kept all properties constant across the dataset and set it up for most linguistic analysis without sacrificing critical contextual and structural information.

\subsection{Exploratory Analysis}

We structured our exploratory analysis into three steps to understand the dataset comprehensively. First, we conducted a linguistic analysis of posts in both the bipolar and non-bipolar groups, comparing structural and grammatical features to identify meaningful language patterns. Second, we explored sentiment variance across the entire dataset to see whether posts between groups have perceivable emotional patterns, thereby quantifying group separation. Finally, we performed a judgmental evaluation on a stratified sample of the posts to verify annotation quality and label reliability for downstream modeling and analysis.

\subsubsection{Linguistic Analysis}

\begin{table}[ht]
\centering
\caption{Linguistic Differences Between Bipolar and Non-Bipolar Posts}
\label{tab:linguistic_features}
\begin{tabular}{|l|c|c|}
\hline
\textbf{Feature} & \textbf{Bipolar} & \textbf{Non-Bipolar} \\
\hline
Average hashtags per post & 0.039 & 0.230 \\
Average URLs per post     & 0.014 & 0.153 \\
Average post length (characters) & 894.36 & 959.17 \\
Average tokens per post   & 80.83 & 93.06 \\
\hline
Average nouns per post    & 33.80 & 42.81 \\
Average verbs per post    & 38.68 & 34.28 \\
Average adjectives per post & 14.76 & 13.78 \\
Average pronouns per post & 24.18 & 19.00 \\
\hline
\end{tabular}
\end{table}

To better understand the linguistic patterns of users, we compared several structural and grammatical elements in bipolar and non-bipolar labeled posts as displayed in Table~\ref{tab:linguistic_features}. Non-bipolar posts contained significantly more external references: 0.230 hashtags and 0.153 URLs per post (vs. 0.039 and 0.014 in bipolar posts). This suggests that non-bipolar users more frequently share news/links or engage in broader conversations, while bipolar posts tend to be introspective. Non-bipolar posts were also slightly longer (959 characters and 93 tokens on average vs. 894 characters and 81 tokens for bipolar posts).

Analysis of part-of-speech (POS) tagging also revealed variations in language styles. Bipolar posts had more action and self-referential writing, reflected in the higher average number of verbs (38.68 vs. 34.28) and pronouns (24.18 vs. 19.00) per post. This suggests a trajectory towards personal storytelling and emotional disclosure. Non-bipolar posts had more nouns (42.81 vs. 33.80), reflecting a more descriptive or topic-driven communication style. Adjective usage was also slightly higher in bipolar posts (14.76 vs 13.78), possibly indicating more expressive or emotionally intense language. These linguistic differences complement the behavioral difference between the two groups and validate the underlying assumptions of our classification task.

\subsubsection{Sentiment Variance Analysis}
\label{subsec:sentiment_variance}

\begin{figure}[ht]
    \centering
    \includegraphics[width=0.9\linewidth]{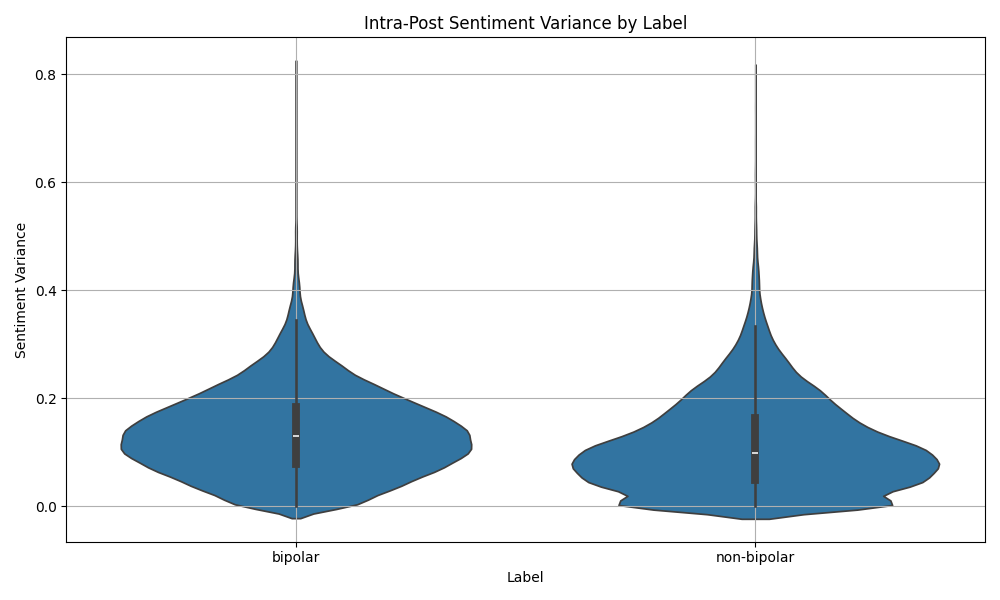}
    \caption{Distribution of intra-post sentiment variance for bipolar and non-bipolar posts}
    \label{fig:sentiment_variance}
\end{figure}

\textbf{}Figure~\ref{fig:sentiment_variance} shows the intra-post sentiment variance distribution of bipolar and non-bipolar posts using a violin plot. While there is some overlap between the two, the bipolar posts are more spread out and have a higher median variance. This trend suggests that posts labeled as bipolar tend to have larger emotional fluctuations within a single post, consistent with clinical features of bipolar disorder, which often include mood swings, depressive episodes, manic symptoms, and anxiety. 

To statistically verify this observation, we conducted a Mann–Whitney U test~\cite{mcknight2010mann}, which indicated an extremely significant difference between the two distributions ($U = 373,025,861.5$, $p < 1e\textsuperscript{-280}$). 

Together, the visual (violin plots) and statistical evidence corroborate the hypothesis that bipolar ideation is related to greater intra-post sentiment variation. The results also validate our annotation policy by linking linguistic features in the posts to typical bipolar expression patterns.

\subsubsection{Judgmental Analysis}
\label{subsec:judgmental_analysis}

To further validate the correctness of our dataset labeling, we supplemented semantic variance analysis with human judgment, following established practices~\cite{hasan2021survey, Landis1977}. We randomly selected 1006 posts (approximately 2\% of the full dataset), stratified across annotated classes, for manual tagging by two annotators using pre-defined rules. Posts were labeled as bipolar if they exhibited emotional fluctuations (e.g., rapid shifts in energy, impulsivity, or depressive loops) and as non-bipolar otherwise. The non-bipolar class included both non-bipolar mental health posts (e.g., r/depression, r/anxiety) and control groups (e.g., r/confidence, r/geopolitics). Annotator concordance yielded Cohen’s Kappa scores of 0.83 and 0.875, indicating ``almost perfect agreement'' ($\kappa > 0.8$), confirming labeling reliability~\cite{Landis1977}.

\subsection{Classification Models and Embedding Strategies}

We evaluated two model categories for detecting posts indicative of bipolar disorder:  

\begin{enumerate}
    \item Transformer-based models: We fine-tuned these pre-trained language models on our bipolar dataset. In particular, we experimented with BERT-base-uncased, RoBERTa-base, DistilBERT, ALBERT, and ELECTRA---a diverse collection of transformer architectures with varying sizes and training corpora.  BERT is a 12-layer bidirectional transformer, initially trained on general English text; RoBERTa is an aggressively optimized version of BERT with more training data; DistilBERT is a compressed (distilled) BERT with fewer parameters; ALBERT is a lighter model parameterizing fewer parameters using factorized embedding parametrization and repetition; ELECTRA employs a discriminator-generator framework for pre-training with fewer samples. We fine-tuned each model in a typical text classification configuration: a [CLS] token encoding (or equivalent) is fed through a feed-forward layer to output the binary label. 

    \item LSTM-based models: These sequential models employ LSTM networks to process posts. We tested both unidirectional and bidirectional LSTM (BiLSTM), which processes sequences forward and backward, to generate fixed-dimensional post-representations. A dense classification layer was added atop the LSTM/BiLSTM outputs. Additionally, we examined an attention mechanism applied to these outputs. In the attention variant, instead of taking a single final hidden state, we employed a learned attention weighting across all the time-step outputs and took a weighted sum as the post-representation. This allows the model to potentially attend to significant regions of the post (e.g., a particular sentence where the user's tone indicates manic thinking). We considered attention as an add-on component and compared LSTM vs. LSTM+Attention scenarios. Significantly, for the LSTM models, we experimented with various embedding inputs:
    
        \begin{itemize}
            \item BERT embeddings: We utilized the pre-trained BERT model as a feature extractor, where for each token in a post, we received a contextual embedding from BERT (i.e., we took the last hidden layer of an off-the-shelf pre-trained BERT-base-uncased model, but not fine-tuned on our task). The token embeddings extracted from the last hidden state of the BERT pretrained model served as input features to the LSTM/BiLSTM. The LSTM was then trained on predicting the sequence from these dense contextual vectors. We froze BERT in this setup (no gradient through to BERT itself), so this tests a hybrid solution: a fixed contextual embedding layer with a trainable LSTM classifier on top.
            
            \item GloVe embeddings: We used 300-dimensional GloVe word vectors (pre-trained on Common Crawl, 840B tokens) as fixed embeddings. Each word in the post is translated to its GloVe vector, and the vectors form the input sequence for the LSTM.
            
            \item Word2Vec embeddings: We employed 300-dimensional Word2Vec vectors (e.g., the Google News corpus vectors) as an alternative static embedding option. These are older vectors and trained on news and web pages. The integration is identical to the one with GloVe above.
    \end{itemize} 

    We set the dimensionality of the LSTM hidden state to 128 units and a dropout rate of 0.2 for the unidirectional LSTM (i.e., 256 total in the BiLSTM, 128 per direction) to keep the model as small as possible. In the attention variant, the attention layer had one feed-forward layer that produced weights over each time step's hidden state.

\end{enumerate}

\section{Experiments and Results}

We employed 5-fold stratified cross-validation for a rigorous and unbiased model assessment. The whole dataset of $\sim$49K posts was divided strategically through stratified sampling: 80\% for training and validation, and the remaining 20\% as a hold-out test set. This provides five independent dataset splits, so five versions of every model are trained. The final label for each test sample was obtained by integrating predictions from the five cross-validation folds. The most frequent class label across these five models was taken as the final prediction. This ensemble approach helps to decrease individual model bias and improve overall prediction stability. This cross-validation approach enables a comprehensive assessment of model performance across different training setups, increasing the robustness of our evaluation framework.

Hyperparameter tuning was performed using Ray Tune\footnote{https://www.ray.io/}, which automates grid search across possible hyperparameter values to determine optimal settings. After an extensive search, the highest-scoring models all shared the same learning rate of $10^{-6}$ and weight decay of $10^{-2}$, resulting in strong performance across most tested architectures. We employed the AdamW optimizer with a learning rate scheduler set to a patience of five epochs~\cite{loshchilov2018decoupled}. This would only alter the learning rate if performance plateaus, avoiding potential unnecessary oscillations during training.

Models were implemented using the PyTorch framework and HuggingFace library, leveraging their rich NLP tools. Our computing infrastructure consisted of two NVIDIA Ampere A100 GPUs with 6912 CUDA cores and 40GB of RAM each, and an additional 512 GB of system memory. This setup was well-equipped to train complex transformer-based models.

\begin{table*}[htb]
    \centering
    \begin{threeparttable}
        \caption{Performance Evaluation of Transformer and LSTM-based Models}
        \label{tab:performance_evaluation}
        \begin{tabular}{|c|c||cccc||c|}
            \hline
            \textbf{Embedding} & \textbf{Model} & \textbf{Accuracy (\%)} & \textbf{F1 (\%)} & \textbf{Precision (\%)} & \textbf{Recall (\%)} & \textbf{Time (hrs)\tnote{*}} \\  
            \hline

            RoBERTa & RoBERTa & 98.13 & 98.08 & 98.38 & 97.79 & 1.20 \\
            BERT & BERT & 98.09 & 98.05 & 98.15 & 97.95 & 1.65 \\
            ALBERT & ALBERT & 97.99 & 97.95 & 98.18 & 97.71 & 1.13 \\
            DistilBERT & DistilBERT & 97.86 & 97.82 & 97.96 & 97.67 & 0.81 \\
            ELECTRA & ELECTRA & 98.10 & 98.06 & 98.04 & 98.08 & 1.50 \\
            \hline
            BERT & BiLSTM + Attention & 98.16 & 98.12 & 98.48 & 97.77 & 2.43 \\
            BERT & BiLSTM & 98.10 & 98.06 & 98.28 & 97.84 & 2.13 \\
            BERT & LSTM + Attention & 98.15 & 98.10 & 98.37 & 97.84 & 2.38 \\
            BERT & LSTM & 98.04 & 98.00 & 98.10 & 97.90 & 2.53 \\
            \hline
            Glove & BiLSTM + Attention & 50.98 & 0 & 0 & 0 & - \\
            Glove & BiLSTM & 50.98 & 0 & 0 & 0 & - \\
            Glove & LSTM + Attention & 51.0 & 0.07 & 100.0 & 0.04 & - \\
            Glove & LSTM & 50.98 & 0 & 0 & 0 & - \\
            \hline
            Word2Vec & BiLSTM + Attention & 50.99 & 0.04 & 100.0 & 0.02 & - \\
            Word2Vec & BiLSTM & 50.98 & 0 & 0 & 0 & - \\
            Word2Vec & LSTM + Attention & 50.98 & 0 & 0 & 0 & - \\
            Word2Vec & LSTM & 50.98 & 0 & 0 & 0 & - \\
            \hline
        \end{tabular}

        \begin{tablenotes}[flushleft]
            \centering
            \item[*] Approx. training time to achieve a best-performing model; this may vary depending on hardware specifications.
        \end{tablenotes}
    \end{threeparttable}

\end{table*}

Table~\ref{tab:performance_evaluation} compares model performance in four evaluation metrics: accuracy, F1 score, precision, and recall. For completeness, we also include an estimate of training time to produce the best-performing model for each architecture to indicate the computational complexity of the different approaches.

\subsection{How do transformer-based models perform in bipolar ideation detection compared to traditional LSTM-based architectures?}

In this study, transformer models like RoBERTa, BERT, and their lightweight versions like DistilBERT emerged as reliable candidates to identify bipolar ideation from social media content. Of these, RoBERTa performed exceedingly well, over 98\% in F1 score and accuracy. These results suggest that transformer models, especially those trained on large and diverse datasets, are well-suited to identify the subtle, often suggestive words used while talking about mental conditions and bipolar disorder.

Interestingly, models with older architectures like LSTM and BiLSTM also performed well, but only when paired with high-quality embeddings. When we gave these models contextualized representation from BERT, their performance was surprisingly close to that of the transformer models.

One possible reason for the existing divergence is how the two systems approach context. Transformers can inherently process the entire input in one pass and assign weights between all words, whereas LSTMs iterate step-by-step and are at risk of forgetting prior context in longer entries. That capacity to view further context in a single pass gives transformers an advantage when dealing with emotionally dense or loosely structured posts.

All else being equal, transformer models are still the way to go unless computational resources are a concern. But in resource-limited settings, well-implemented LSTM models with contextualized embeddings can still be a valid and competitive choice for detecting bipolar disorder on social media.

\subsection{To what extent does embedding selection (contextual vs. static) influence model performance?}

Our evaluation suggests that the word embedding nature plays a larger role in determining classification performance in bipolar disorder detection than model architecture. Traditional sequence models like BiLSTM and LSTM, when paired with context-aware embeddings from BERT, yielded remarkably high scores. For example, the attention-based BiLSTM model achieved an F1 score of 98.12\%, even surpassing RoBERTa, which achieved 98.08\%. To our surprise, even without adding attention mechanisms, LSTM models with BERT embeddings consistently crossed the 98\% F1 and accuracy thresholds. These findings highlight that most models' success comes from the semantic density packed in the input, not because of model complexity.

In contrast, the same recurrent architectures performed more or less uncompetitively when static embeddings such as GloVe or Word2Vec were employed. The F1 scores achieved as a consequence varied around zero, and the models were biased towards majority class predictions. This sudden decline is because of the intrinsic limitations of static embeddings that give a fixed representation to every word regardless of context. These representations fall short in handling the kind of subtle, emotionally charged, and figurative language prevalent in bipolar and mental health speech.

Collectively, these findings lead us to think that contextual depth and quality of embeddings are more important than classifier design at the underlying level. Architectural tweaks like bidirectionality or attention can bring marginal gains, but can never perform well for poor input representations. On the other hand, if given good-quality contextual embeddings, relatively lightweight models can catch up with and surpass heavier models. This is especially worth it in settings where computational resources are scarce. For instance, A light LSTM paired with a BERT embedding might be a more realistic choice than a fully fine-tuned transformer model in a resource-constrained environment.

Finally, this result is consistent with trends seen in other NLP tasks within the mental health field: situating language is of critical importance. A model's success or failure may very well come down not as much to how it processes data, but to the quality of data concerning meaning capture. Thus, future work in this field could be assisted by further emphasis on embedding methods, such as improved methods for extracting and adapting contextual representations.
\subsection{Does adding attention mechanisms to LSTM/BiLSTM architectures improve performance significantly?}

Integrating attention mechanisms slightly enhances LSTM-based models, especially when context embeddings such as BERT are used. For instance, a BiLSTM using BERT embeddings had an F1 score of 98.06\%, and the inclusion of an attention mechanism boosted it slightly to 98.12\%. Although the improvement is minimal ($\sim$0.06\%), this shows that attention allows the model to attend to the most important pieces of the sequence.

Based on our evaluation, attention mechanisms are helpful but a secondary matter compared to quality embedding in analyzing clinical text. In higher-scoring configurations with contextual embeddings, attention is helpful but not essential. It may have some marginal benefit in lower-scoring configurations, but it cannot compensate for embedding shortfalls.

\subsection{How does training time vary across different models and embedding strategies, and what are the practical trade-offs?}

Training time also greatly varies between models and needs to be a strong consideration in real deployment. In transformer models, DistilBERT stands out particularly in its efficiency, requiring only 0.81 hours per fold but producing a competitive F1 score of 97.82\%. In comparison, the training time for RoBERTa is 1.20 hours, while for BERT it stands at 1.65 hours, with barely a significant improvement over DistilBERT in F1 and accuracy metrics.

LSTM-based models with BERT embeddings took longer to train; for example, LSTM required 2.53 hours per fold, since the overhead of retrieving contextual embeddings per token occurs dynamically during training. This time can be accelerated by pre-computing and caching the embeddings, although with added complexity in training.

Static embedding models didn't offer viable training times but are expected to take longer to converge. Due to less informative input features, they often need more epochs.

In summary, DistilBERT offers the best trade-off between accuracy and training efficiency. Nonetheless, hybrid BERT-LSTM models can be a choice in systems where transformer fine-tuning and training competence are optional.

\section{Conclusion and Future Work}

In this study, we compared the performance of several transformer-based models with different LSTM architectures for bipolar disorder detection using Reddit posts. Based on our experimental results, we summarize the key findings as follows: 

\begin{enumerate}
    \item BiLSTM + Attention using BERT embeddings performed the best (98.12\% F1 score) followed closely by RoBERTa (98.08\% F1 score).
    
    \item Among the transformer models, RoBERTa was the best, although all transformer models were close, with comparable results (above 97.8\% F1 score), showing themselves capable of this classification task.
    
    \item The lightweight DistilBERT model required nearly half the training time of larger transformers like BERT and RoBERTa, making it an ideal choice for efficient training on large mental health datasets.
    
    \item BiLSTM and LSTM models with frozen BERT embeddings performed comparably and in some cases slightly outperformed fine-tuned transformer models.

    \item Static embeddings like GloVe and Word2Vec failed to enable learning as hoped and are proof of the effectiveness of contextualized representations in mental health text analysis.
\end{enumerate}

Our future research will expand the analysis across cross-platform datasets, potentially enabling the models to generalize across different user groups. We also plan to explore multimodal and ensemble methods, combining high-performing models to boost robustness. Finally, we intend to integrate explainability techniques to enhance transparency and support clinical relevance in real-world deployment scenarios.

\bibliographystyle{IEEEtran}
\bibliography{references}

\end{document}